# A Historical Interaction between Artificial Intelligence and Philosophy

Youheng Zhang

School of Marxism, Wuhan University of Technology, Wuhan 430070, China

zhangyouheng@whut.edu.cn

**Abstract:** This paper reviews the historical development of AI and representative philosophical thinking from the perspective of the research paradigm. Additionally, it considers the methodology and applications of AI from a philosophical perspective and anticipates its continued advancement. In the history of AI, Symbolism and connectionism are the two main paradigms in AI research. Symbolism holds that the world can be explained by symbols and dealt with through precise, logical processes, but connectionism believes this process should be implemented through artificial neural networks. Regardless of how intelligent machines or programs should achieve their smart goals, the historical development of AI demonstrates the best answer at this time. Still, it is not the final answer of AI research.

**Keywords**: connectionism; symbolism; AI; the history of AI; the philosophy of AI

## 1. Introduction

Even in mythological times, humans aspired to create intelligent machines. The ancient Egyptians devised a "shortcut" for themselves by building statues that priests could hide while delivering wise guidance to the populace. This "scam" has continued to occur throughout the development of AI. The idea of artificial intelligence originates in philosophy, logic, and mathematics, and it is now a reality. In the fourth century B.C., Aristotle pioneered data abstraction. His formal logic provides a framework for competent scientific reasoning and a foundation for further research. His difference between matter and form is still one of the essential principles in computer science today. Data abstraction is the separation of concepts from their actual representations or procedures (forms) from the shells of encapsulated methods. G. Leibniz, a 17th-century philosopher, significantly impacted modern algebra, algorithms, and symbolic logic. He believed that symbols might be utilized to express how people think. Leibniz's work influenced G. Boole, a 19th-century mathematician. In his book [1], Boole describes an essential method of symbolic reasoning and claims to process logical propositions with arbitrary terms in pure notation to make reasonable, logical inferences. To show intelligence, computers must be able to reason; that's where "Boolean algebra" comes in. A. Turing, a computer scientist, published a paper in a philosophical journal in the twentieth century [2]. This paper's publication is regarded as "Setting Sail" for AI. It describes the famous Turing test and speculates the possibility of programming intelligence into computers [3] (p.56). The Dartmouth Conference's organizer, J. McCarthy, introduced the concrete concept of AI in 1956, putting the science of giving computers "intelligence" into the public's consciousness for the first time.

As of now, many questions remain in the field of AI. The public feels AI progress is based on hype rather than genuine advancement. ANNs (Artificial Neural Networks), or connectionism[1], the basis of this enthusiasm in the AI profession, raises concerns about how far AI can progress. Since Turing initially proposed the concept of intelligent machines, there have been two primary philosophies in AI research: symbolism, which had significant success in the 20th century, and connectionism, which reached its apex in the 21st century. In contrast to symbolism, which prioritized logical thinking, ANNs were first motivated by biology and intended to imitate the functionality of biological neural networks with computer neural networks to enable AI to perform intelligent activities. Connectionist AI has achieved groundbreaking advances in recent research and has found extensive applications. It is even to argue that the concept of ANNs was a "Copernican Revolution" in an era where symbolic thinking predominated. In contrast to the serial method of symbolic AI, the parallel program of ANNs allows AI to do simple commonsense tasks such as face recognition, speech recognition, unmanned driving, and so on. However, the "neurons" replicated on the virtual machine are qualitatively different from the actual neurons in the brain. The connectionist design of artificial neural networks cannot be comparable to biological neural networks. Furthermore, according to the definition of strong AI, the ANNs can't reach this aim genuinely because of the absence of a logical structure.

## 2. Historical Development of AI

In terms of how to provide the program "intelligence," the symbolic research approach was first supported by the majority in the field. Symbolic AI progressed in two stages from the 1950s to the 1990s: ATP (Automatic Theorem Proving) and ES (Expert System). Although symbolism "lost its luster" after the 1990s, with the advent of "Knowledge Graph" technology in the early 21st century, the discipline of KR (Knowledge Representation) based on symbolism has received increased attention. Unlike the path of Symbolism from prevalence to decline, the development of ANNs is more "ups and downs." From the 1940s to the present: the early development of connectionist AI can be divided into two major stages: "Perceptron" and BP (Back Propagation); with the wave of the Internet in the 1990s, AI research once ceased until the proposal of DL (Deep Learning) in 2006, and connectionism ushered in its wave once more.

### 2.1. Historical Development of Symbolism AI
#### 2.1.1. 1950-1960 Automatic Theorem Proving's Initial Research

ATP is the intellectual origin and theoretical foundation of symbolic AI [4]. It originated in logic and was designed primarily to automate logical calculations. Martin Davis, a logician, created the first ATP program in 1954 and, later in his career, collaborated with the philosopher Putnam to create the widely known D-P (Davis-Putnam) algorithm in 1960 and the later upgraded DPLL algorithm [5] [6]. However, the "Logical Theorist" program [7], completed in 1956 by A. Newell and H. Simon, gained more attention and is regarded as one of the most important original works in the history of AI. This program proves a large portion of the propositional logic section of the first volume of Russell's *Principia Mathematica*. In 1957, D. Prawitz created a programming language to implement his natural deduction algorithm. IBM recruited Paul Gelernter to implement a geometric theorem prover as early as 1956, and he turned to logician Paul Gilmore in 1957. However, Gilmore first achieved the Semantic Tableau method, which is analogous to Prawitz's. Grant then gave the Geometry Machine program in 1959 [8]. In the same year, the logician Hao Wang achieved even greater success in the field of ATP: he proved the 150 first-order logic and 200 propositions of the Principia Mathematica on an IBM 704 computer. ATP laid a solid and dependable foundation for the subsequent development of the symbolic AI research agenda. Many scholars and research groups, such as Wenjun Wu in China and the Argonne Group in the US, have significantly contributed to the ATP development since then. After 1960, symbolic AI applications completely dominated the historical stage.

#### 2.1.2. 1960-1990 Expert System's Prime Period

Due to the widespread adoption of ES, the symbolist paradigm enjoyed significant success in the early years of research. ES has many characteristics that make it appropriate for AI research and development, including separating the knowledge base from the reasoning machine. DENDRAL was the first ES program to apply symbolic reasoning to complex sciences using heuristics. Its program contained a wealth of scientific knowledge in the field of chemistry [9]. E. Feigenbaum, the program's creator, was a student of H. Simon, and his collaborator was geneticist J. Lederberg. The shared interest in the philosophy of science of the latter facilitated their long and fruitful collaboration. In 1965, the duo enlisted the help of chemist C. Djerassi to launch this interdisciplinary research program at Stanford [10], hoping that DENDRAL would serve as an "intelligent assistant" to chemists, performing many tasks like an expert but not necessarily at the level of an expert in theory. One of the DENDRAL participants, philosopher B. Buchanan, wrote in a paper with Feigenbaum:

One of the early motivations for our work was to use heuristic search to reason in complex scientific problems, and the other was to use AI methods better to understand some fundamental questions in the philosophy of science. [11]

Under Buchanan's guidance, Ph.D. E. Shortliife, a biomedical scientist, created the expert system MYCIN [12], which like DENDRAL, was written in the LISP language to help prescribe drugs to blood-borne patients. However, MYCIN was never used in clinical trials, and its successor, EMYCIN, was developed and marketed by Van Melle in 1981 [13]. According to Newell, MYCIN is the pioneer of expert systems because it created the generative rule that is a component of expert systems: approximate reasoning. MYCIN and the thousands of expert systems that followed it demonstrated the power of a small amount of knowledge to enable intelligent decision-making procedures in many vital areas during the 1980s [3] (p.59). However, after the failure of FGCS (Japan's Fifth Generation Computer) in 1988, there are now few companies in the market dedicated to developing ES, but the ES is still an essential part of AI research and has significant relevance to the field of Knowledge Representation.

### 2.1.3. 1970-Present Knowledge Representation's Undisturbed Progress

ES and NLP (Natural Language Processing) spawned KR [14]. Humans need a lot of knowledge to perform intelligent activities, and AI needs, too. It can perform intellectual tasks by structuring knowledge using symbolic techniques like frames, logic, word vectors, scripts, and semantic networks. The main mathematical and logical foundations of knowledge representation are: (1) In 1974, M. L. Minsky presented "Frame Theory," which led to an object-oriented design philosophy that influenced the design languages of related programs [15]; (2) In 1975, Roger Schank issued "Script Theory" in which scripts organize the knowledge base in the form of frames [16];(3) In 1976, John Sowa proposed the "Conceptual Graph," which is a knowledge representation system based on Semantic Network and Peirce's logic [17]; (4) In 1981, Rudolf Wille introduced FCA (Formal Concept Analysis), which is a fundamental method for deriving a conceptual hierarchy or formal ontology from a set of objects and their attributes [18]. In 1981, the United States launched the CYC project to develop an encyclopedia (i.e., the first Knowledge Graph) in response to Japan's FGCS project. D. Lenat and R. Guha, the project's main organizers, were respectively Buchanan, McCarthy, and Feigenbaum students. Under their leadership, the CYC project sought to build an AI system that was neither narrow nor fragile, Lenat had predicted that CYC would need 10 million properly organized pieces of information, including rules and facts, and the CYC team tried to take into account all empirical knowledge [19]. But in 1985, The Wordnet project, started in 1985 by G. Miller, was used more practically by "creating a machine-readable database linked by lexical and semantic associations" [20]; it did not have the rules that CYC does. In 2001, T. Berners-Lee introduced the Semantic Web, which is dependent on the World Wide Web for its use but differs from the W3C in that it is based on a machine-readable database provided by the Web [21]. Other databases that emphasize machine readability include DBpedia, Yago, Metaweb, etc. But METAWEB's Metaweb, which Google later acquired in 2010 and incorporated into Freebase. Google released its Knowledge Graph in 2015 and shut down Freebase in 2016, moving all of its databases to Wikidate.

## 2.2. Historical Development of Connectionism AI
### 2.2.1. 1940-1970 From ANN's Invention to Perception's Frustration

Before Turing became the "ghost myth" of AI, another thought trend, cybernetics, also gave stimulus to the birth of AI. Norbert Wiener founded this interdisciplinary field from the standpoint of science philosophy. This discipline invariably influenced early ANNs studies and researchers, who modeled AI from a biological, communication, and engineering perspective rather than logic and mathematics. The followers of Wiener, W. Pitts, and W. McCulloch proposed the M-P model in their 1943 paper [22], which was based on the properties of biological neural networks: In the presence of inhibitory input, the neuron cannot be activated by stimulus input (output=0); in the absence of inhibitory input, the neuron is activated if the sum of the inputs exceeds a predetermined threshold (output=1). They attempted to record the behavior of complicated nets in the notation of the symbolic logic of propositions starting connectionism AI up. In 1957, under the influence of the Hebb-Learning rule[2], psychologist Frank Rosenblatt created an artificial neural network model called Perception [23], which can handle some basic visual tasks. The M-P model's absolute inhibition rule no longer holds in the Perceptron because each input is weighted, the input weights are no longer uniform, and some stimulus inputs have inhibitory effects. Because of Rosenblatt's fame, some American media even declared that "machines are about to simulate human consciousness." In 1963, Rosenblatt published a book summarizing his research, but it was immediately criticized by M. L. Minsky [24], who believed in symbolism. Minsky and Seymour Papert pointed out that Perceptual machines could not solve the XOR problem [25], demonstrating that the limited nature of ANNs research at the time was fatally flawed, resulting in the loss of nearly a decade of ANNs research funding.

### 2.2.2. 1980-1990 From Hopefield Network's Stimulation to Back Propagation's Struggle

BP (Back Propagation), a new algorithm created in 1974 by P. Werbos, was influenced by Freud's psychoanalytic theory. According to the Ph.D. thesis of Werbos, BP adds an extra layer (hidden layer) to the neural network, thus solving the XOR problem [26]. But because ANNs research was at its lowest point, little attention was paid to his work. However, the first revival of ANNs did not come from the field of AI itself but rather from the physicist J. Hopfield. His work's outcomes, known as Hopfield Networks, it physics-based and influenced by spin systems [27]. A set of neurons make up this multi-loop feedback system, in which there is the same number of feedback loops as neurons, and each neuron only performs output to other neurons besides itself. The Hopfield Network not only resolves the XOR problem but also shows its usefulness by exploring approximations to combinatorial optimization and NP-complete problems[3] [28]. This sparked a new wave of ANNs research; after the publication of PDP (Parallel Distributed Processing), a compilation of papers, in 1986, a group of still-optimistic researchers returned to the field. This momentum in cognitive and computer science proposes that the brain may be modeled as a parallel processing system, where individual neural units are linked, and activation signals flow across this network, with the entire process occurring concurrently and dispersed. When referring to connectionism in AI, it is to PDP [29]. One of the participants in this movement, G. Hinton, brought the BP algorithm back to prominence with his cooperative paper [30]; this algorithm iteratively adjusts the weights of the connections in the network to minimize the difference between the network's actual output vector and the desired output vector. But then, the BP algorithm only applied to networks with a small number of hidden layers. When the connectionists used the BP algorithm to multilayer neural networks, the BP network became a "slow learner." A multilayer network could take millions of iterations to learn, resulting in increasingly limited success. The Internet's onslaught gradually drowned out the connectionists' voices in the 1990s.

### 2.2.3. 2006-Present From Deep Learning's Resurgence to the Complex ANNs

The Internet and AI technology have overlapping goals, and the faster and easier interconnection also hints at connectionism. The massive amount of Internet-generated data gives neural networks more significant opportunities. In 2006, Hinton's article 31 in *Science* lifted the curtain; this paper describes a machine learning method for converting high-dimensional data into low-dimensional code and then pre-training it layer by layer. DL (Deep Learning), a field mentioned in the previous century, has gained utility via Hinton's work. DL is a machine learning method that differs from others in that the trainer no longer has to label the dataset; instead, the machine actively learns the dataset's features. So far, we can find two main areas of ANN research: the method (algorithm) and neural network construction. In the past century, numerous neural network models have been hypothesized or presented as half-baked. Still, there have been no effective algorithms to improve their application, and algorithms were the "right-hand man" to advance these models. Early in its development, AI drew inspiration from various fields, and after absorbing much, AI has broken through their framework and started to influence them in turn [3] (p.56). The revival of DL research has resulted in a renewed emphasis on some artificial neural networks: (1) CNNs (Convolutional Neural Networks) are neural networks that are primarily used in machine vision. The "convolutional layer" is its kernel, which transforms a function into another function, the "pooling layer" selects the main features of the object, and the "fully connected layer" gives the probability that the object is something; (2) RNNs (Recurrent Neural Networks) are neural networks that are primarily used for natural language processing. It is characterized by the emphasis on sequential data or time series. It converts independent activation of hidden layers into dependent activation, forming a cycle to obtain a specific memory capacity. (3) GNNs (Graph Neural Networks) are a type of artificial neural network that is not conventional. It is concerned with analytic graphs of objects, in which nodes represent entities and edges represent relationships between them. Such graphical connections can be used for single or sets of objects.; (4) QNNs (Quantum Neural Networks) are advanced neural networks designed for quantum computers. It applies quantum theory to the study of ANNs, intending to overcome some of the limitations of traditional ANNs. The macroscopic goal of neural networks like these is to simulate brain intelligence, which reminds us of Rosenblatt's dream, and isn't that exactly what he wished for?

## 3. Philosophical Thinking in AI Research

Whether symbolic or connectionist, the initial design of an AI program comes from a theoretical proposition rather than from each precise computation step. Each AI worker's intelligent proposition suggests a practical solution for machine intelligence based on their research, which is already veering toward philosophical thinking. As a result, some AI experts, as well as philosophers, have thought about the philosophy of AI.

### 3.1 Turing

"We might hope that machines would eventually compete with humans in all areas of pure intelligence" [2] (p.460), as Turing stated in his essay on the philosophy of AI. His ideas paved the way for later AI study. This paper's foundation role and contribution to AI are as follows: (I)It introduces the "Imitation Game," which asserts that electronic computers can learn to think from their functional aspects. In the face of theological, mathematical, neuroscience, and computer science refutation, Turing demonstrates the possibility of machine thinking by disproving it [2] (pp.443-454); (II) It makes the case that punishment plays a part in the machine learning process and that AI can learn active judgment through reward and punishment signals by using non-emotional communication [2] (p.457) ;(III) Explain the intelligent system's unpredictable behavior. Intelligent behavior may deviate slightly from completely self-consistent behavior in the calculation, but it will not result in random behavior or meaningless repeated cycles[2] (p.459); (IV) Turing's learning machine is instructive for modern machine learning. His view of the "accidental deviation of computer intelligent behavior"[2] (p.459) mentioned in this article is still helpful for training and learning algorithms today. Turing's vision may hardly have come true, but he left the question, "Where is the best place to begin research? ...chess or hardware?" [2] (p.460) initiated the tug of war in the AI era.

### 3.2 *Computer Science as Empirical Inquiry: Symbols and Search*

The paper [32] by the symbolism's founders, Simon and Newell, includes philosophical reflections on the symbolic paradigm. The symbolic AI agenda first clarifies the type of research that will be done: "We wish to speak of computer science as empirical research." [32] (p.114). This qualitative statement is significant because it expresses the fundamental notion of the "intelligent" design of symbolic AI and the fundamental mindset toward the scientific study of AI, which is not a metaphysical fantasy but rather an empirical simulation and logical inference of known human intelligence. The creation of machines and programs allows for analyzing existing phenomena and discovering new ones [32] (p.114). Secondly, it brings attention to the link between intelligence and symbols:

The fundamental functions of intelligence cannot be explained by any "intelligence principle," just as there is no "life principle" to describe what life is all about. The capacity to store and analyze symbols is essential for intelligence. [32] (p.115)

AI science theory and practice both make efforts about what "intelligence" is and how to achieve it. The underlying premise of symbolism is that "intelligent activity requires a physical symbol system, and intelligent behavior can be carried out by a physical symbol system" [32] (p.118). It is easy to deduce the following idea if we adopt this stance: "Man is characterized by a physical symbol system" [32] (p.119). Symbolic systems use a heuristic search procedure to solve problems, which brings us to the last point about how intelligence is achieved. An organized problem-solving method is what makes up a heuristic algorithm, and this behavioral process represents intelligence. The development of a comprehensive problem domain in any setting does, in fact, show intelligence. But this "problem domain" is closed as opposed to open in symbolic AI; therefore, it is done on purpose, and this view is based on the idea that "symbolic systems do not demonstrate intelligence when they are in a chaotic condition" [32] (p.126).

### 3.3 *A Logical Calculus of the Ideas Immanent in Nervous Activity*

Pitts and McCulloch certainly had no idea their influence would continue for nearly a century. Their cooperation paper [22] was an opening work of the philosophy of connectionist AI; McCulloch blended philosophical theory and practice and tested his philosophical views through Pitts's research. "The law of 'all-or-none' neural activity can characterize any neuron's activity as a proposition" [22] (p.117). This groundbreaking theory reduces the brain and mind to a mathematical model, paving the path for philosophers to describe the mind in new ways. "The physiological relationships between brain activity relate to propositional relations" [22] (p.117). Logical relations between propositions describe the causal flow of brain activity. McCulloch and Pitts appealed to this logical relation in time, making each process of neural activity definite and thus revealing the "veil" of neural activity. They proposed two conjectures based on the theory above: the "nets without circle" and the "nets with circle" theory. The "nets without circle," a neural network in which synapses do not form loops, is known as the "M-P model," which can be implemented by this non-recursive neural network, i.e., it executes the function of Turing machines; the class of "nets with circle" is a neural network with circular connections. The acyclic networks are more complex than those for acyclic networks because the time of neural activity in the loop is unknown, requiring numerous quantizations. Later, S. Kleene improved the theory of "nets with circle" by identifying patterns in them by expressing the set of all input neuron activation sequences as mathematical expressions that lead a given network with a cyclic network to a particular state following complete processing [33]. Although the logical language used by McCulloch and Pitts in this article is cumbersome and complex, the concept of modeling the brain from mathematics illustrates the mind and mind-body relationship from a different perspective; it's a significant victory for computationalism over traditional philosophical incursions.

### 3.4 About Philosophical Insights of AI Philosophy

Vladimir Lenin once said intelligent idealism is better than stupid materialism[4], and the same holds for AI philosophy: an intelligent philosopher is better than a silly expert, and an intelligent expert is better than a foolish philosopher. The philosophy of AI can be traced back to Turing. However, it is not limited to him. Phenomenologist Hubert Dreyfus made a name for himself with his critique of AI, including the biological, psychological, epistemological, and ontological hypotheses. He deemed that "the last three hypotheses are not empirical but philosophical and thus can be critiqued philosophically" [34] (p.69). Dreyfus was a "layman" with no formal training in computers. Still, his impact cannot be explained solely by social considerations because his work successfully unites philosophy with AI, and more similar work is being done in the present.

Dreyfus claims that "the human brain may process information in a completely different way than a digital computer" [34] (p.72) and criticize ANNs on a biological level. Dreyfus contends that the M-P neural model only partially describes how neurons function. Given humans' limited experience, ANNs are just an uncritical explanation adopted by neuroscientists and AI professionals. Still, AI specialists take this as the official explanation. His criticism utilized philosophical synthesis to challenge AI's analysis, yet philosophical synthesis was founded on concrete analysis. Dreyfus' straightforward analysis incurs widespread criticism in philosophy and the AI field. Even Dreyfus' position is changing. He and his cousins[5] revisited their criticism in the 1980s with the decline of symbolism and the revival of ANNs. He has certain expectations for ANNs, and in the articles, they jointly published, the scope of AI criticism has been reduced to symbolism.

Dreyfus' critique of symbolic AI can be boiled down to the three primary "philosophical hypotheses. The psychological hypothesis critique poses the issue, "Can the mind work like a computer, i.e., is it reasonable to employ a psychological computer model?" [34] (p.75). According to this theory, human thinking is hierarchical, and computers can replicate human thinking by accessing it from an appropriate level known as the "information processing level." Dreyfus opposed this view and thought that "The term information is utilized in this theory in a specific sense, not to be confused with its ordinary meaning and implications" [34] (p.77). Therefore, even though a computer can process information heuristically, it cannot be assumed that a computer has mental activity and the brain's mental activity is a program. The epistemic hypothesis critique has a flaw in that "AI experts believe that all non-arbitrary behavior can be defined according to specific rules, and whatever those rules are, computers can always be used to replicate those non-arbitrary norms" [34] (p.102). Dreyfus argues that there are no perfect laws of conduct and distinguishes between neurological and physicochemical principles, arguing that "meaningful human action cannot be viewed simply as the physical movement of the human organism." [34] (p.106). He also draws on linguistic arguments that machines produce arbitrary interpretations according to formal rules, i.e., that they cannot understand semantics, and questions the symbols' loss of the essence of the formalized object. The issue with the hypothetical ontological critique is that intelligent acts must be seen as separate determined elements in principle [34] (p.118). Dreyfus here does not investigate the machine as a subject (though there is a propensity to do so); instead, he attacks the inclusion of intelligence in the symbolic processing process. He deemed that "The world must be expressed as itself a structured set of descriptions consisting of starting elements, and this is an essential prerequisite for all AI research activity" [34] (pp.123-124). Consequently, symbolic ontology's harshest problem shouldn't formalize and model intelligence.

Dreyfus asserted that "The valid methods that philosophy finds in the natural sciences must also be valid in AI research" [35]; however, he used nebulous language in traditional philosophy to support his position, claiming that some computer languages and AI programming have long been mentioned in conventional philosophy and that this absurd method of proof has received the same criticism. Unlike Dreyfus, philosopher of science H. Putnam was a booster in the early development of ATP and a pioneer in some aspects of AI. Some of his philosophical thought is frequently accepted by insiders. He developed a functionalist approach to the philosophy of mind by making an analogy between the functions of the mind and those of computers, suggesting that the state of the human mind is also a process of input, processing, and output. Putnam began his book [36] with two well-known thought experiments, "brain in a vat" and "twin earth," both aimed at rejecting the phenomenological concept of "Intentionality." Putnam contended that "Comprehension is not defined by internal mental phenomena, but by the subject's ability to use them and produce the proper mental phenomena in the right context" [36] (pp.19-20), so that "Understanding of external things do not necessitate a series of mental phenomena" [36] (p.20). AI struggles to understand "Intentionality" because "Intentionality" is not "Intentional."The current progress of AI, such as adding new references to the understanding of machines, the ontological connection in the Knowledge Graph, and the logical structure in the artificial nervous system, all demonstrate the effectiveness of Putnam's functionalism argument. The development of AI technology has not been halted by the concept of Intentionality in traditional philosophy.

## 4. Rethinking AI Research from a Philosophical Perspective
### 4.1 The Real World is Not Exactly Like Symbols.

The advancement of computational capacity on hardware favors the current development of AI, and the ATP, the keystone of symbolic AI, illustrates that it is difficult to see the dawn even though the hardware is rich. The easy difficulties are solved by 80 percent at the beginning, while the rest is difficult to break through. The significance of ATP is that it solves mathematical logic issues, and the importance of solving problems for AI is the ability to cognize objects. Cognition involves symbolic processing, and symbolic processors have the same cognitive powers as humans, drawing philosophers' attention to AI [37] (p.200). Mathematics and logic are fantastic "metaphysics" that construct thought's superstructure, and mathematical logic forms the expressions of events. One strategy for AGI (Artificial General Intelligence) is to use mathematical logic to formalize common sense information so that common sense problems can be solved using logical reasoning [38]. To attain the AGI, ES is vital, and Knowledge Representation can be considered a route. Although it is theoretically plausible to create an all-around ES, in practical, this takes immense databases in practice to mimic the human expert. It assumes that human experts base their judgment on various conditional information while making decisions. Each expert already has a rule collection in their brain, and all the expert system architect has to do is extract the rules and program them into the computer [39]. Even so, the reasoning is not a necessary and sufficient condition for someone or some system to be an expert; a skilled car driver reacts instinctively rather than reasoning when confronted with an unexpected situation, and a chef does not have to debate whether to wash or chop vegetables first to ensure taste when working. Although nonmonotonic reasoning can help to solve a portion of the problem and bring the machine's inference conclusions closer to reality, computer scientists may be overly idealistic in their approach, treating the world of common sense as the world of science. Even in science, there are so many uncertainties and contingencies that one must proceed cautiously by trial and error. The mode of things is not as idealized in the realm of common sense as in the world of science; mistakes and random events are common. Much specialized work has been performed an umpteen times before becoming a human expert. Once a machine has reasoned, it encounters concerns that might no less than a human expert. Feigenbaum also wrote in his book:

Expert systems do not yet understand these things. Part of learning to be an expert is to understand not only the letter of the rules but also their spirit...... He knows when to break the rules and what is relevant to his task and irrelevant to his job.[40]

### 4.2 The Connectionism AI Requires a Logical Framework.

IBM's Deep Blue defeated G. Kasparov in 1997, which utilized Brute-force search, and DeepMind's Alpha Star defeated a human champion playing Starcraft Ⅱ in 2019. In several computer games, connectionist AI has outperformed human players due to neural networks trained using supervised learning and reinforcement learning. However, Alpha and the different Game AI can only be employed in specified situations. Alpha Go, which plays Go, can't play Starcraft, not only because the training of the neural networks is unique but also because these networks make decisions rather than reasoning[6]. This weight-based decision making is regarded as measurement, whereas reasoning is viewed as logic, and both capacities exist in the human brain. Even though the two abilities are somewhat similar, i.e., both are sound judgments made in a context. However, the ability to make decisions in an environment with clear rules and goals is not the same as the ability to make decisions in an open environment, which is why connectionism AI cannot be used as a trader in a stock market with clear rules but an open environment. One of the most significant limitations is that synaptic remodeling, neurotransmitter, and hormone neuromodulation on neuronal excitability at the micro level cannot achieve large-scale synchronous spiking activity and global connection at the macro level [41]. Connectionism's network models can still not imitate the brain at multiple levels. To overcome the limits of ANI (Artificial Narrow Intelligence), "systems that combine the expressiveness and programmatic diversity of symbolic systems with the ambiguity and adaptability of connectionist expressions must be developed"[42].

Research on NeSy(Neural-Symbolic) AI, a synthesis of connectionism and symbolism, has been advancing theoretically, but it has been challenging to make significant progress in real-world applications. The reasons are that (1) we only have a limited understanding of the underlying mechanisms, even though the human brain can solve highly abstract reasoning problems using entirely physical neural networks, and (2) since the human brain is made up of neurons that work by exchanging chattering bioelectrical impulses rather than symbols like words [43] (p.467), the way logical structures in thinking work is different from how the human brain functions. Despite these significant disparities, P. Blazek and M. Lin's recent study has taken the neurosymbolic issue a step further：

We develop a simple yet scalable neurocognitive model of neural information processing in which each neuron is a specialized decision-making agent within a hierarchical network. To demonstrate that this model is sufficient to encode complex cognitive capabilities, we implemented it in a general-purpose machine learning algorithm for classification tasks that build deep neural networks that are explainable, capable of symbolic manipulation, and better reflect the neurobiological properties of the brain. These networks perform well on standard AI benchmarks and surpass existing deep neural networks on tasks involving reasoning and outof-distribution generalization. [44] (p.607)

To some extent, ENNs (Essence Neural Networks) have overcome the "unpredictability" caused by backpropagation and stochastic gradient descent. ENNs establish logical structures by the absolute distinction of Concept neurons and relative distinction of Differentia neurons to simulate symbolic AI program functions, in which Differentia neurons, Subconcept neurons, and Concept neurons all have different levels of distinctive parts. "ENNs establish a comprehensive computational foundation for creating AGI based on interpreting cognitive neurons," according to Blazek [44] (p.607). ENNs have put a lot of effort into developing logical reasoning, but it is still insufficient, and the training process is still a congenital condition for the neurons' intelligence. While ENNs may be able to translate texts and recognize images well, these abilities still fall short of AGI. This at least is a good start, and AI programs may not necessarily need to mimic human cognitive processes directly. But it's still possible to reflect on the structure of human cognition through understanding AI cognitive processes and thereby learn how we think and how our brains function contribute to that thinking. One of AI's functions is to give our understanding of human thought more structure; if thinking is viewed as merely one thing, it would be impossible to delve deeper and comprehend more [45].

### 4.3 The Distinction Between Artificial and Biological Intelligence

Despite evident parallels between human reasoning and computer program execution, we lack the basic understanding of implementation in biological and artificial neural networks [43] (p.467). The human brain is frequently equated to a computer because current AI systems outperform any other known artificial machine performing complicated tasks. The human brain has often been compared to the most technologically advanced machine throughout history. Examples include Descartes' comparison of the human brain to a hydraulic press in the 16th century and Freud's steam engine-inspired psychoanalysis studies in the 19th century. If some artificial machine surpasses AI in the future, this metaphor may only be restricted by human imagination or technological progress. However, even though the two can operate similarly in function, they may have quite different internal workings. AI is still simply a mimic of BI (Biological Intelligence).

Learning is a capacity that could be used to show intelligence, but in bio and AI, the learning process of AI is very different from how humans learn, even though Alpha Star has figured out how to win at video games. There is an a priori structure of cognition in the biological brain before any learning takes place, which German philosopher Immanuel Kant referred to as the " Synthetic A Priori Judgments "[7]. Natural talents such as sense of time, space, mathematical knowledge, avoidance of danger, and so on are partially genetically derived, and it's the byproduct of biological evolution. They are not acquired through learning at first, but rather BI brought about by instinct. This structure is configured with vital information that allows us to acquire knowledge. Through learning, it is possible to continuously make adjustments to the original construction of the knowledge representation and to make the newly acquired knowledge and abilities appear consistent and structured in the brain; this structure for AI is typically trained from a dataset, and the weights in the artificial neurons are adjusted by algorithmic optimization. For AI, the form of the knowledge representation is empirical rather than a priori, and it does not change over time. Usually, it is feasible to alter the AI's learning objectives so that the weights in the neurons adjust once more. However, the AI program can still not learn various skills or knowledge.

This demonstrates that learning is a comprehensive behavior in BI, whereas a specialized behavior in AI. As a result, while AI can match or surpass human capabilities in some areas of learning and function, it is still unable to reach the comprehensive level of intelligence, which includes the abilities to "comprehend by analogy" and "make the connection." This type of programming for AI functionalization may face a potential obstacle:

Studying the brain at an empirical level and limiting it to the level of precise "programming," i.e., the level of "logic gates" in computers; however, the brain is at the biological level of neurons, and even if all the pulse-linked sequences of neurons are written, is this sufficient to account for "neural computation"? [46]

Without a doubt, no. It would be exceedingly challenging to come up with an explanation that would make the biological rules governing brain activity and the algorithmic rules governing thinking coherently, even if we could explore the brain more deeply in the future. AI and BI differ not only in the hierarchy but also in nature. AI must be able to choose the best programs to address new issues that arise in various environments if they want to adapt to the real world and become the real intelligent. Meta-program is a program that can perform a selection process. AI programming may be more efficient than BI in executing a specific task, but this program does not and cannot function as a meta-program; on the contrary, only BI can serve as a meta-program for all biological behavior and mental. Life is the vehicle for such a meta-program generated from the historical process of survival and reproduction of living beings, while an AI program written does not and will never have this unique nature. Unless there is a subversive cultural and ideological revolution to abolish the concepts like "life," there will always be an "intelligence" distinction between AI and BI.

### 4.4 Analyzing the Development of AI Via the Lens of Philosophical History

AI is a concretization of philosophical ideas; on the one hand, it aids in answering and explaining some conventional philosophical concerns; on the other hand, philosophy advances this goal. The history of computer science demonstrates that philosophy is far more sophisticated than smattering assumed and that it will always be able to offer new and exciting contributions to the discipline [37] (p. 195). Looking at the symbolism from the beginning of AI research, this top-down strategy can be referred to as "think what comes," while connectionism's bottom-up approach can be "take what you can." The former "mesis" human intelligence while the latter "Methexis"[8]. From Aristotle and Plato to medieval nominalism and realism to the current argument between empiricism and rationalism alone, the conflict between them has been a recurring theme throughout the history of philosophy. The history of the theory of thought provides clues about how AI will develop. The development of AI as a philosophical idea come true is still essentially a result of this evolution. Considering the FGCS led to the dream of ES bidding farewell and using logic to the extreme in AI research did not show that the FGCS was more effective; instead, it was more like Hume bringing empiricism to the grave and putting symbolism to an end. But the development of thought theories will never come to the finale. Like the argument over AI between analytical philosophy and phenomenology, where one side takes a scientific-analytical approach to the issue, and the other rejects the use of cognizant scientific consciousness to explain consciousness itself. The opposing viewpoints won't stop as long as AI research is conducted, but through reflecting on the development of philosophy, they will finally come together. NeSy AI will merge symbolism and connectionism as Kant fused past philosophy and paved a new way for future philosophy.

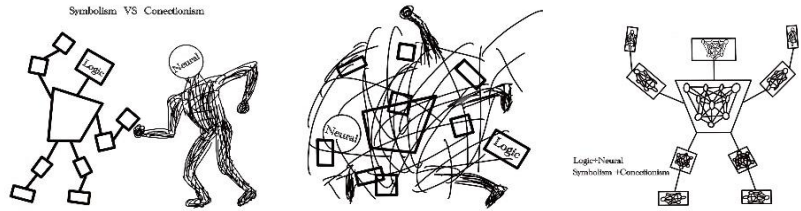

Figure1. Conflict and merging of theories.

NeSy is not novel; instead, it is a natural consequence of the advancement of AI theory. It combines the dogmas of "think what comes" and "take what you can." Its beginnings can be ascribed to the essay [22] of McCulloch and Pitts in 1943, whose research results act as a vital inspiration for Norbert Wiener's Cybernetics. Wiener and Turing are the "Plato" and "Aristotle" of AI history, and their contributions to connectionism and symbolism are epic. The development of AI promoted from their main ideas is analogous to the evolution of Plato's and Aristotle's essential ideas in the history of philosophy. Although their theoretical constructs and hypotheses do not clearly indicate the progression and goals of research development, this cannot slow the flow of AI technology from moving forward, "the progress of philosophy lies in breaking through its limitations without jettisoning clarity, and so does the progress of AI" [37] (p. 201). As a result, NeSy as "Kant" will not be the final model of AI, but it will be developed through the "Fichte, Schelling, and Hegel"[9] processes. Previous research work programs and indications of limited success hint at the undiscovered and some methodologies and mechanisms that can be applied to construct actual AI [3] (p.60). The "Hegelian" model won't be the final form of AI; somewhat, it will resemble the development of the history of philosophy. This completion model will still create something new, just like the various philosophical schools after Hegel, each of which concerns some common and distinctive fields. Therefore the ANI will develop into AGI and subsequently API (artificial professional intelligence) based on it in somedays. This demonstrates how philosophy influences AI from a historical and methodological perspective ([18], p-537); hence, the history of technology in AI is the philosophical history of the metaphysical transformation into reality.

## 5. Conclusions

Looking back at the history of technology and philosophy, we can see that philosophy and AI have always affected each other. Philosophy discovers new questions due to AI's exploration, and AI responds to these questions as it evolves. Philosophy and AI have grown profoundly intertwined, resulting in the philosophy of AI, which numerous academic institutions have acknowledged. Engineers with a strong philosophical background find it easier to express the aims and paths of AI research, and AI offers a potent and practical tool for philosophical contemplation. Additionally, understanding the philosophy of AI necessitates not only a philosophical foundation and familiarity with the AI field but also a historical review of this interdisciplinary field. To make this discipline can benefit all more than just "malarky," it will need to involve the united efforts of experts in philosophy, AI, and the history of science and technology. To prematurely declare the invalidity and failure of a research paradigm is to belittle scientific inquiry, just as Minsky previously proclaimed the end of ANNs and it returned in its progress. Scientific inquiry is decentralized, requiring numerous researchers to build entire "nests" like "ants" to explore new areas. The symbolic paradigm is undeniably no longer mainstream in modern times, and ANNs publications and various rankings have dominated mainstream AI research attention. The "Symbolic Regression" appears to be motivated by the necessity of learning algorithms, but it also highlights the significance of symbolic approaches.

Even though ANNs have made significant progress recently, connectionism AI is still not sufficiently "intelligent." Not because artificial neural networks differ from biological networks, but because the problems the AI developers choose to solve determine how intelligently those networks are designed. If AI wants to achieve its goal of being intelligent, it must make its own decisions and look for ways to solve problems. Although AI technology is intended to improve human wellbeing rather than to serve as a goal in and of itself, there is no assurance that this will always be the case. Many AI scholars have envisioned strong AI as the "grail" of the discipline. Still, this goal is so lofty or challenging that many AI professionals have chosen to remain silent to focus on more specific issues. Looking back in history, this phenomenon could have its reason. If we can go back to the Middle Ages and gather the best philosophers of the time to discuss how to design and build an intelligent machine. They might come up with something brilliant, but it would never be the "right answer" today because the concept of AI would not emerge until centuries later; similarly, today's experts in AI would not appear until centuries later. They are already outstanding masters, but they have yet to find the perfect concept for AI. However, the advancement of AI still relies on human ponder, and experience may not always provide the right solutions. Hence the purpose of AI philosophy research is never merely to pose questions and criticize the achievement of AI but rather to identify some practical, theoretical cornerstone for AI research through philosophical analysis and critical thought.

## Note

[1] The term originates from psychology and cognitive science and refers to a methodology in AI.
[2] According to Herb's learning rule, as the brain processes new information, neurons become active and linked together to form neural networks. These connections are initially weak, but with continued stimulation, they become stronger and stronger, and as a result, the weights between them rise.
[3] The traveling salesman problem, a well-known combinatorial optimization issue, is one of the most understandable examples of an NP-complete problem. The best route with the lowest cost for the salesman must be determined given a collection of cities, their distances, and the costs associated with traveling between them while beginning and ending in the same location and making only one visit.
[4] Which can see in *Lenin Collected Works: Volume 38: Philosophical Notebooks*. https://www.marxists.org/archive/lenin/works/cw/volume38.htm (accessed 2022-07-24).
[5] Stuart Dreyfus, whose research interests are in artificial neural networks, is the relative of Hubert Dreyfus.
[6] The Starcraft game map is a coordinate system like the Go. The difference is that resources, the environment, and other elements must be considered to win the game. Still, both contain identical basic rules, such as an attack, defense, layout, and other fundamental game techniques.
[7] According to Kant's viewpoint, the a priori structure of cognition logically precedes empirical knowledge and causes it to occur.
[8] Mimesis, which means to mimic, and methexis, which refers to the relationship between a particular and a form, are two of Plato's philosophical categories. Aristotle pointed out the contradiction between these categories.
[9] J. Fichte, F. Schelling, and G. Hegel are the three German classical philosophers who made significant contributions after I. Kant

## References

1. Boole, G. *An Investigation of the Laws of Thought: On Which Are Founded the Mathematical Theories of Logic and Probabilities*; Cambridge Library Collection - Mathematics; Cambridge University Press: Cambridge, 2009. https://doi.org/10.1017/CBO9780511693090.
2. Turing, A. M. Computing Machinery and Intelligence. *Mind* **1950**, LIX, 433–460. https://doi.org/10.1093/mind/LIX.236.433.
3. Buchanan, B. G. A (Very) Brief History of Artificial Intelligence. *AI Magazine* **2005**, 26, 53–60. https://doi.org/10.1609/aimag.v26i4.1848.
4. Pastre, D. Automated Theorem Proving in Mathematics. *Ann Math Artif Intell* **1993**, *8*, 425–447. https://doi.org/10.1007/BF01530801.
5. Davis, M.; Putnam, H. A Computing Procedure for Quantification Theory. *J. ACM* **1960**, *7*, 201–215. https://doi.org/10.1145/321033.321034.
6. Davis, M.; Logemann, G.; Loveland, D. A Machine Program for Theorem-Proving. *Commun. ACM* **1962**, *5*, 394–397. https://doi.org/10.1145/368273.368557.
7. Newell, A.; Simon, H. A. The Logic Theory Machine. A Complex Information Processing System. *Journal of Symbolic Logic* **1957**, *22*, 331–332. https://doi.org/10.2307/2963663.
8. Wang, H. Toward Mechanical Mathematics. *IBM Journal of Research and Development* **1960**, *4*, 2–22. https://doi.org/10.1147/rd.41.0002.

9.  Lederberg, J. How DENDRAL Was Conceived and Born. In *A history of medical informatics*; Bruce, B., Karen, D., Eds.; Association for Computing Machinery: New York, NY, USA, 1990; pp 14–44.
10. Lederberg, J.; Sutherland, G. L.; Buchanan, B. G.; Feigenbaum, E. A.; Robertson, A. V.; Duffield, A. M.; Djerassi, C. Applications of Artificial Intelligence for Chemical Inference. I. Number of Possible Organic Compounds. Acyclic Structures Containing Carbon, Hydrogen, Oxygen, and Nitrogen. *J. Am. Chem. Soc.* **1969**, *91* (11), 2973–2976. https://doi.org/10.1021/ja01039a025.
11. Buchanan, B. G.; Feigenbaum, E. A. Dendral and Meta-Dendral: Their Applications Dimension. *Artificial Intelligence* **1978**, *11* (1–2), 5–24. https://doi.org/10.1016/0004-3702(78)90010-3.
12. Shortlife, E. H. The Computer as Clinical Consultant. *Archives of Internal Medicine* **1980**, *140* (3), 313–314. https://doi.org/10.1001/archinte.1980.00330150027008.v.
13. Van Melle, W.; Shortliffe, E. H.; Buchanan, B. G. EMYCIN: A Knowledge Engineer's Tool for Constructing Rule-Based Expert Systems. In *Rule-based expert systems*; Addison-Wesley: Boston, MA, USA, 1984; pp 302–313.
14. Weischedel, R. M. Knowledge Representation and Natural Language Processing. *Proc. IEEE* **1986**, *74* (7), 905–920. https://doi.org/10.1109/PROC.1986.13571.
15. Bayle, A. J. Frames: A Heuristic Critical Review. In *Eighth Annual International Phoenix Conference on Computers and Communications. 1989 Conference Proceedings*; IEEE Comput. Soc. Press: Scottsdale, AZ, USA, **1989**; pp 624–628. https://doi.org/10.1109/PCCC.1989.37457.
16. Schank, R. C.; Abelson, R. P. Scripts, Plans, and Knowledge. In *Proceedings of the 4th International Joint Conference on Artificial Intelligence - Volume 1*; IJCAI'75; Morgan Kaufmann Publishers Inc.: San Francisco, CA, USA, **1975**; pp 151–157. https://doi.org/DOI: 10.5555/1624626.1624649.
17. Sowa, J. F. Conceptual Graphs for a Data Base Interface. *IBM J. Res. & Dev.* **1976**, *20* (4), 336–357. https://doi.org/10.1147/rd.204.0336.
18. Wille, R. Restructuring Lattice Theory: An Approach Based on Hierarchies of Concepts. In *Ordered Sets*; Rival, I., Eds.; Springer Netherlands: Dordrecht, **1982**; pp 445–470. https://doi.org/10.1007/978-94-009-7798-3_15.
19. Elkan, C.; Greiner, R. Building Large Knowledge-Based Systems: Representation and Inference in the Cyc Project: D.B. Lenat and R.V. Guha. *Artificial Intelligence* **1993**, *61* (1), 41–52. https://doi.org/10.1016/0004-3702(93)90092-P.
20. Miller, G. A. WordNet: A Lexical Database for English. *Commun. ACM* **1995**, *38* (11), 39–41. https://doi.org/10.1145/219717.219748.
21. Berners-Lee, T.; Hendler, J.; Lassila, O. The Semantic Web. *Scientific American* **2001**, *284* (5), 34–43. https://doi.org/10.1038/scientificamerican0501-34.
22. McCulloch, W. S.; Pitts, W. A Logical Calculus of the Ideas Immanent in Nervous Activity. *The bulletin of mathematical biophysics* **1943**, *5* (4), 115–133. https://doi.org/10.1007/BF02478259.
23. Rosenblatt, F. The Perceptron: A Probabilistic Model for Information Storage and Organization in the Brain. *Psychological Review* **1958**, *65* (6), 386–408. https://doi.org/10.1037/h0042519.
24. Van Der Malsburg, C. Frank Rosenblatt: Principles of Neurodynamics: Perceptrons and the Theory of Brain Mechanisms. In *Brain Theory*; Palm, G., Aertsen, A., Eds.; Springer: Berlin, Heidelberg, **1986**; pp 245–248. https://doi.org/10.1007/978-3-642-70911-1_20.
25. Huang, G.-B. What Are Extreme Learning Machines? Filling the Gap Between Frank Rosenblatt's Dream and John von Neumann's Puzzle. *Cognitive Computation* **2015**, *7* (3), 263–278. https://doi.org/10.1007/s12559-015-9333-0.
26. WERBOS, P. Beyond Regression: "New Tools for Prediction and Analysis in the Behavioral Sciences. Ph. D. dissertation, Harvard University 1974.
27. Hopfield, J. J. Neural Networks and Physical Systems with Emergent Collective Computational Abilities. *Proceedings of the National Academy of Sciences* **1982**, *79* (8), 2554–2558. https://doi.org/10.1073/pnas.79.8.2554.
28. Hopfield, J. J.; Tank, D. W. "Neural" Computation of Decisions in Optimization Problems. *Biol. Cybern.* **1985**, *52* (3), 141–152. https://doi.org/10.1007/BF00339943.
29. Todd, P. M. A Connectionist Approach to Algorithmic Composition. *Computer Music Journal* **1989**, *13* (4), 27–43. https://doi.org/10.2307/3679551.
30. Rumelhart, D. E.; Hinton, G. E.; Williams, R. J. Learning Representations by Back-Propagating Errors. *Nature* **1986**, *323* (6088), 533–536. https://doi.org/10.1038/323533a0.
31. Hinton, G. E.; Salakhutdinov, R. R. Reducing the Dimensionality of Data with Neural Networks. *Science* **2006**, *313* (5786), 504–507. https://doi.org/10.1126/science.1127647.
32. Newell, A.; Simon, H. A. Computer Science as Empirical Inquiry: Symbols and Search. *Commun. ACM* **1976**, *19* (3), 113–126. https://doi.org/10.1145/360018.360022.
33. Kleene, S. C. Representation of Events in Nerve Nets and Finite Automata. In *Automata Studies. (AM-34), Volume 34*; Shannon, C. E., McCarthy, J., Eds.; Princeton University Press, 2016; pp 3–42. https://doi.org/doi:10.1515/9781400882618-002.
34. Dreyfus, H. L. *What Computers Can't Do: The Limits of Artificial Intelligence*, Revised, Subsequent edition.; HarperCollins: New York, NY, USA, 1978.
35. Dreyfus, H. L.; Dreyfus, S. E. Making a Mind Versus Modelling the Brain: Artificial Intelligence Back at the Branchpoint. In *Understanding the Artificial: On the Future Shape of Artificial Intelligence*; Negrotti, M., Ed.; Springer London: London, 1991; pp 33–54. https://doi.org/10.1007/978-1-4471-1776-6_3.

36. Putnam, H. *Reason, Truth and History*; Cambridge University Press, 1981. https://doi.org/10.1017/CBO9780511625398.
37. Glymour, C. Artificial Intelligence Is Philosophy. In *Aspects of Artificial Intelligence*; Fetzer, J. H., Ed.; Springer Netherlands: Dordrecht, 1988; pp 195–207. https://doi.org/10.1007/978-94-009-2699-8_7.
38. McCarthy, J. Artificial Intelligence, Logic and Formalizing Common Sense. In *Philosophical Logic and Artificial Intelligence*; Thomason, R. H., Ed.; Springer Netherlands: Dordrecht, 1989; pp 161–190. https://doi.org/10.1007/978-94-009-2448-2_6.
39. Dreyfus, H. L.; Dreyfus, S. E. From Socrates to Expert Systems: The Limits of Calculative Rationality. *Technology in Society* **1984**, *6* (3), 217–233. https://doi.org/10.1016/0160-791X(84)90034-4.
40. Feigenbaum, E.; McCorduck, P. *The Fifth Generation: Artificial Intelligence and Japan's Computer Challenge to the World*; Addison-Wesley Longman Publishing Co., Inc.: USA, 1983.
41. Macpherson, T.; Churchland, A.; Sejnowski, T.; DiCarlo, J.; Kamitani, Y.; Takahashi, H.; Hikida, T. Natural and Artificial Intelligence: A Brief Introduction to the Interplay between AI and Neuroscience Research. *Neural Networks* **2021**, *144*, 603–613. https://doi.org/10.1016/j.neunet.2021.09.018.
42. Minsky, M. L. Logical Versus Analogical or Symbolic Versus Connectionist or Neat Versus Scruffy. *AI Magazine* **1991**, *12* (2), 34-51. https://doi.org/10.1609/aimag.v12i2.894.
43. Jaeger, H. Deep Neural Reasoning. *Nature* **2016**, *538* (7626), 467–468. https://doi.org/10.1038/nature19477.
44. Blazek, P. J.; Lin, M. M. Explainable Neural Networks That Simulate Reasoning. *Nature Computational Science* **2021**, *1* (9), 607–618. https://doi.org/10.1038/s43588-021-00132-w.
45. Pagels, H. R.; Dreyfus, H. L.; Mccarthy, J.; Minsky, M. L.; Papert, S.; Searle, J. Has Artificial Intelligence Research Illuminated Human Thinking? *Annals of the New York Academy of Sciences* **1984**, *426*, 138–160. https://doi.org/10.1111/j.1749-6632.1984.tb16517.x.
46. Bell, A. J. Levels and Loops: The Future of Artificial Intelligence and Neuroscience. *Philos Trans R Soc Lond B Biol Sci* **1999**, *354* (1392), 2013–2020. https://doi.org/10.1098/rstb.1999.0540.
47. Schiaffonati, V. A Framework for the Foundation of the Philosophy of Artificial Intelligence. *Minds and Machines* **2003**, *13* (4), 537–552. https://doi.org/10.1023/A:1026252817929.